\documentclass[11pt]{article}

\usepackage[final]{acl}

\usepackage[utf8]{inputenc}
\usepackage[T1]{fontenc}
\usepackage{times}
\usepackage{latexsym}
\usepackage{booktabs}
\usepackage{multirow}
\usepackage{amsmath}
\usepackage{amsfonts}
\usepackage{microtype}
\usepackage{inconsolata}
\usepackage{xcolor}
\usepackage{graphicx}
\usepackage{float}
\usepackage{pifont}
\usepackage[most]{tcolorbox}

\definecolor{claimback}{HTML}{F7FAFC}
\definecolor{claimrule}{HTML}{2F5EA8}
\definecolor{claimtext}{HTML}{1F2937}
\newtcolorbox{claimcallout}{
  enhanced,
  breakable,
  colback=claimback,
  colframe=claimrule!35,
  boxrule=0.35pt,
  leftrule=2.8pt,
  arc=2pt,
  outer arc=2pt,
  left=8pt,
  right=8pt,
  top=6pt,
  bottom=6pt,
  boxsep=0pt,
  before skip=8pt,
  after skip=8pt
}

\makeatletter
\newcommand{\authornote}[2]{%
  \g@addto@macro\@thanks{\footnotetext[#1]{#2}}%
}
\makeatother

\title{Beyond GRPO and On-Policy Distillation: \\
An Empirical Sparse-to-Dense Reward Principle for Language-Model Post-Training}

\author{
  Hejian Sang\textsuperscript{1}\Thanks{Equal contribution. Author order was determined randomly.} \and
  Yuanda Xu\textsuperscript{1}\footnotemark[1]\footnotemark[4] \and
  Zhengze Zhou\textsuperscript{1}\footnotemark[1] \\
  \bfseries Ran He\textsuperscript{1}\footnotemark[1] \and
  Zhipeng Wang\textsuperscript{2}\footnotemark[2] \and
  Alborz Geramifard\textsuperscript{1}\footnotemark[3] \\
  \normalfont\textsuperscript{1}LinkedIn \quad \textsuperscript{2}Google
  \authornote{2}{\texttt{jasonzwang@google.com}}%
  \authornote{3}{\texttt{agf@linkedin.com}}%
  \authornote{4}{Corresponding author: \texttt{ericxu@linkedin.com}}%
}

\begin{document}

\maketitle

\begin{abstract}
In settings where labeled verifiable training data is the binding constraint, each checked example should be allocated to the model and reward density where it is most informative. We identify a reward-density principle: sparse sequence-level reward is most useful on models that can explore and discover better behavior, while dense token-level teacher supervision is better suited for compressing that behavior into a smaller deployment model; accordingly, scarce labeled data should be used upstream on the strongest available teacher, then transferred downstream as dense supervision.

We evaluate this rule through a four-stage workflow---teacher RL, forward-KL warmup, on-policy distillation, optional post-bridge student RL---on verifiable math and code generation with Qwen3, Llama, and Qwen2.5 models. At fixed Qwen3-1.7B deployment-student size, an RL-improved 8B teacher distilled through the dense bridge outperforms direct GRPO on the same student ($79.3\%$ vs.\ $75.9\%$ on MATH; $25.2\%$ vs.\ $19.8\%$ on AIME~2024, avg@16), while transfer from the same teacher \emph{before} RL underperforms. A component ablation confirms that each stage is load-bearing: the RL-improved teacher adds $7.8$ absolute points on MATH over a raw teacher of the same size, the forward-KL warmup adds $1.7$, and on-policy distillation adds $3.3$. The teacher-quality ordering---raw-teacher transfer $<$ direct GRPO $<$ RL-teacher transfer---replicates on Llama-3.1-8B-Instruct with a Llama-3.3-70B-Instruct teacher and on Qwen2.5-7B-Instruct for code generation with a Qwen2.5-32B-Instruct teacher, where the dense bridge improves HumanEval and MBPP by $3.3$ and $3.8$ absolute points, respectively, over direct student GRPO. The operational lesson is to avoid spending scarce labeled data on the least prepared policy: use sparse reward for teacher-side discovery, dense transfer for student compression, and student-side sparse reward only after the bridge. The OPD implementation used in our experiments is available at \url{https://github.com/HJSang/OPSD_OnPolicyDistillation}.
\end{abstract}

\section{Introduction}
\label{sec:intro}

Labeled training data is the bottleneck of LLM post-training. Pretraining text and teacher rollouts can scale with compute; labeled data for verifiable tasks does not scale so easily. Each example needs a problem with a checkable answer and a grader whose errors will not corrupt the reward. The practical question is therefore not which post-training algorithm is best in isolation, but \emph{which model should receive each scarce labeled example, with which density of signal, and in what order}.

The default approach is to train the deployment model directly: if a 1.7B model is the deployment target for MATH, run GRPO on it directly. This paper argues for a different allocation, and for the simple reward-density principle behind it.

\paragraph{The reward-density principle.} Sparse reward first defines a reward-shaped target distribution. Direct GRPO asks the deployment student to discover this target from its own sparse rollouts. The teacher-first route instead uses the labeled examples to train a stronger teacher, then treats that teacher as a dense, autoregressive proxy for the target. The bridge is the projection step: forward KL moves the student onto teacher support; OPD transfers the teacher under student occupancy. Sparse reward is used where exploration is productive, and dense teacher supervision is used where the goal is compression.

\begin{figure*}[t]
  \centering
  \includegraphics[width=1.5\columnwidth]{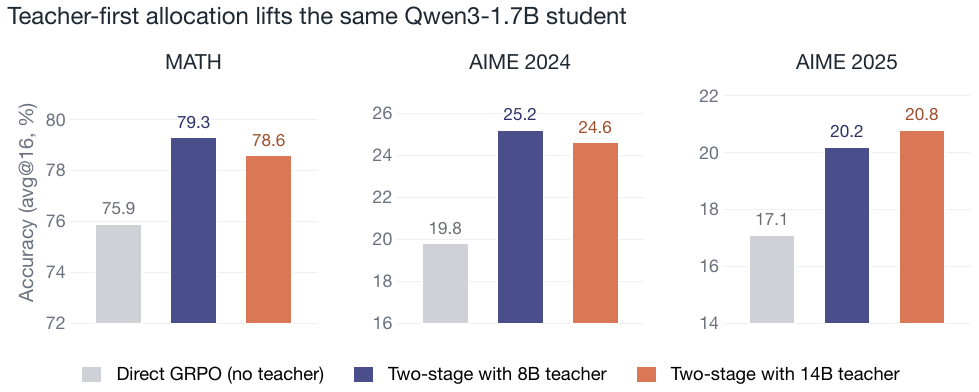}
  \caption{Headline contrast on the same Qwen3-1.7B deployment student (avg@16, \%; details in Table~\ref{tab:qwen-transfer-full}). Each panel uses its own zoomed accuracy axis so the three runs stay visually comparable across MATH and AIME. Allocating the same labeled training data to teacher RL plus the two-stage bridge outperforms direct student GRPO at every benchmark.}
  \label{fig:metrics-summary}
\end{figure*}

\begin{figure*}[t]
  \centering
  \includegraphics[width=0.92\textwidth]{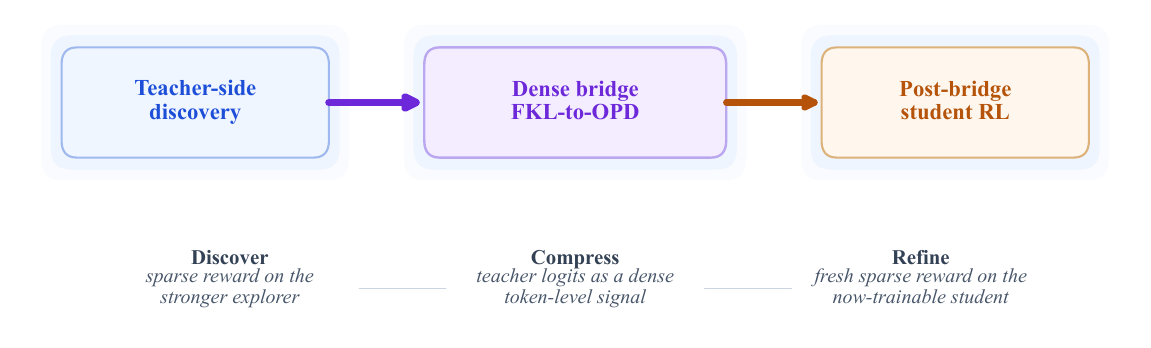}
  \caption{Where labeled training data should be allocated. The teacher-side path (Stage~1: teacher RL) discovers reward-shaped behavior; the two-stage dense bridge (Stage~2a: FKL warmup, Stage~2b: OPD) converts it into token-level supervision for the deployment student; the optional post-bridge student RL stage (Stage~3) uses any remaining labeled data on a now-trainable student.}
  \label{fig:workflow-overview}
\end{figure*}

\paragraph{What this changes in practice.} The standard post-training pipeline---SFT, then RL on the deployment model---places the scarce labeled data in the least effective position first. The teacher-first view prescribes a different order: allocate the labeled training data to a model large enough to use it, run a two-stage dense bridge into the deployment model, and only then decide whether any held-out labeled data remains worth using on the student. Figure~\ref{fig:workflow-overview} summarizes the resulting pipeline.

\paragraph{Contributions.} The individual ingredients of our workflow---teacher GRPO, forward-KL distillation, OPD, and post-bridge student GRPO---are established techniques. Our contribution is the reward-density principle and a controlled study of where scarce labeled data should enter this pipeline. Specifically, we make three contributions:
\begin{enumerate}\setlength\itemsep{2pt}
\item \textbf{Teacher-first allocation} (Section~\ref{sec:experiments}). At fixed deployment-student size, a fixed pool of labeled training data yields a stronger student when it is allocated to teacher RL plus dense transfer than when it is allocated to direct student RL. The gain requires a reward-shaped teacher: transferring the same base teacher before teacher-side RL underperforms direct GRPO, so scale alone is not the cause. The same ordering holds on code generation, where the RL-teacher bridge improves HumanEval and MBPP by $3.3$ and $3.8$ absolute points over direct student GRPO.
\item \textbf{A two-stage dense bridge} (Sections~\ref{sec:theory} and~\ref{sec:exp-ablation}). In the pre-Stage-3 transfer comparison, a forward-KL warmup on teacher rollouts followed by OPD on student rollouts outperforms both teacher-sample SFT and OPD-only transfer. The warmup reduces the cold student's coverage mismatch so that the subsequent OPD stage---which is a local trust-region update under a dense teacher-induced implicit reward---is better conditioned.
\item \textbf{Post-bridge student RL} (Section~\ref{sec:experiments}). The bridge changes student trainability: sparse-reward GRPO that is weak on a cold student lifts the bridge endpoint above both direct GRPO and a matched replay control that reuses bridge data.
\end{enumerate}

\paragraph{Scope.} The evidence spans verifiable math and code generation across three model families (Qwen3, Llama, Qwen2.5); model configurations and evaluation details are in Section~\ref{sec:experiments}.

\paragraph{Terminology.} A \emph{sparse reward} is a sequence-level task reward $R(x,y)$ available only at the end of a trajectory. A \emph{reward-shaped target} $\pi_R^*$ is the KL-regularized policy induced by that reward. A \emph{dense teacher signal} is the token-level teacher log-probability $r_T(s_t,y_t)=\beta\log\pi_T(y_t\mid s_t)$ supplied by a teacher that approximates $\pi_R^*$. \emph{OPD} is reverse-KL distillation on student rollouts. The \emph{two-stage bridge} (or FKL-to-OPD) is forward-KL on teacher rollouts followed by OPD on student rollouts.

\section{The Workflow}
\label{sec:workflow}

We now spell out the operational route implied by the allocation principle. The workflow takes a fixed pool of labeled training data $\mathcal{D}$, a deployment student $\pi_\theta$, and a larger teacher $\pi_T$ that shares the student's tokenizer. It produces a post-trained student through four stages; the theoretical justification is deferred to Section~\ref{sec:theory}.

\paragraph{Stage 1: Teacher-side sparse-reward RL.} Run GRPO (or a comparable verifier-based RL algorithm) on the teacher using $\mathcal{D}$. This produces a reward-shaped teacher $\pi_T$ whose distribution concentrates on high-reward trajectories. Standard recipes either skip this stage or replace it with supervised fine-tuning on $\mathcal{D}$; however, supervised fine-tuning can exhibit incomplete learning even on the training distribution~\citep{xue2026supervised}. We argue in Section~\ref{sec:theory} that an RL-shaped teacher is qualitatively different from an SFT-shaped one as a source of dense supervision.

\paragraph{Stage 2a: Forward-KL warmup on teacher rollouts.} Sample $K$ rollouts from $\pi_T$ on prompts in $\mathcal{D}$ and train the student to match the teacher's next-token distribution on those rollouts:
\[\mathcal{L}_{\mathrm{F}}(\theta) = \mathbb{E}_{s\sim d_{\pi_T}}\operatorname{KL}\!\left(\pi_T(\cdot\mid s)\,\Vert\,\pi_\theta(\cdot\mid s)\right).\]
This is supervised next-token training under teacher occupancy. It is mode-covering, off-policy with respect to the student, and well-conditioned at cold start because it trains directly on teacher-supported states.

\paragraph{Stage 2b: On-policy distillation under student rollouts.} Sample rollouts from the current student $\pi_\theta$ and minimize the reverse-KL to the teacher on those rollouts:
\[\mathcal{L}_{\mathrm{R}}(\theta) = \mathbb{E}_{s\sim d_{\pi_\theta}}\operatorname{KL}\!\left(\pi_\theta(\cdot\mid s)\,\Vert\,\pi_T(\cdot\mid s)\right).\]
The teacher's log-probabilities are queried on-policy at the student's sampled prefixes; the teacher itself is frozen. This stage corrects the student on its own state distribution.

\paragraph{Stage 3 (optional): Post-bridge student-side sparse-reward RL.} If any labeled data remains held out from Stages~1--2, run GRPO on the bridged student using that held-out data. Section~\ref{sec:experiments} shows this stage adds value over (a) skipping it and (b) reusing already-seen bridge data for more student updates.

\begin{claimcallout}
{\small\bfseries\textcolor{claimrule}{The full workflow at a glance}}\par\vspace{2pt}
{\small\color{claimtext}
\textbf{Stage 1.} Sparse-reward RL on teacher, using labeled data $\mathcal{D}$. \\
\textbf{Stage 2a.} Forward-KL distillation on teacher rollouts.\\
\textbf{Stage 2b.} On-policy reverse-KL distillation on student rollouts.\\
\textbf{Stage 3.} (Optional.) Sparse-reward RL on student, using held-out labeled data.\\[2pt]
Stages~1--2 are the core bridge; Stage~3 applies only when labeled data has been split between teacher and student.}
\end{claimcallout}

\paragraph{What standard recipes drop.} The SFT-then-RL recipe on the deployment model corresponds to ``replace Stage~1 with SFT on $\mathcal{D}$, drop Stage~2 entirely, run Stage~3 on the full $\mathcal{D}$.'' DeepSeek-R1-style teacher distillation~\citep{deepseek2025r1} corresponds to ``run Stage~1, keep only the off-policy half of Stage~2 (teacher-sample SFT), drop Stage~2b and Stage~3.'' Standalone on-policy distillation~\citep{agarwal2024onpolicy,lu2025onpolicydistillation} corresponds to ``run Stage~2b alone, dropping Stage~2a.'' The component ablation in Section~\ref{sec:experiments} measures each of these omissions.

\section{Theory: A Reward-Density Principle}
\label{sec:theory}

We justify the workflow through a trust-region reading of on-policy distillation that mirrors recent self-distillation analyses~\citep{shenfeld2026selfdistillation}. The full derivation is in Appendix~\ref{app:opd-derivation}; here we keep the load-bearing equations and arguments.

\paragraph{Reward-shaped target.} Let $x$ be a prompt, $y=(y_1,\ldots,y_T)$ a response, and $s_t=(x,y_{<t})$ the autoregressive state. A sequence-level verifier reward $R(x,y)$ induces the KL-regularized target
\begin{equation}
\pi_R^*(y\mid x)=\tfrac{1}{Z_R(x)}\pi_{\mathrm{ref}}(y\mid x)\exp(R(x,y)/\beta),
\label{eq:reward-shaped-target}
\end{equation}
the optimum of $\mathbb{E}_{y\sim\pi}[R(x,y)]-\beta\operatorname{KL}(\pi\Vert\pi_{\mathrm{ref}})$~\citep{rafailov2024direct,shenfeld2026selfdistillation}. Direct student RL tries to recover $\pi_R^*$ from the student's own sparse rollouts. The workflow instead uses sparse reward on a larger model and treats the resulting teacher as a proxy
\begin{equation}
\pi_T(\cdot\mid x)\approx \pi_R^*(\cdot\mid x).
\label{eq:teacher-proxy-assumption}
\end{equation}

\paragraph{OPD as a local implicit-reward update.} Let $\pi_k=\pi_{\theta_k}$ be the current student. The negative reverse-KL gradient at $\pi_k$ equals the policy gradient of a dense per-token implicit reward $\widetilde R_T^k$:
\begin{equation}
\begin{aligned}
&-\beta\,\nabla_\theta\operatorname{KL}(\pi_\theta\Vert\pi_T)\big|_{\theta_k} \\
&\quad = \mathbb{E}_{y\sim\pi_k}\!\left[\widetilde R_T^k(x,y)\,\nabla_\theta\log\pi_\theta(y\mid x)\right]_{\theta_k}, \\
&\widetilde R_T^k(x,y)=\sum_{t=1}^T\beta\log\tfrac{\pi_T(y_t\mid s_t)}{\pi_k(y_t\mid s_t)}.
\end{aligned}
\label{eq:opd-as-irl}
\end{equation}
Each OPD step is therefore the local trust-region update that maximizes $\mathbb{E}_\pi[\widetilde R_T^k] - \beta\operatorname{KL}(\pi\Vert\pi_k)$. The signal is a dense, on-policy, teacher-induced surrogate for the sparse verifier; its value depends on two conditions on the proxy.

\paragraph{(C1) Optimality.} The teacher must achieve near-maximal reward on the verifier: $\mathbb{E}_{y\sim\pi_T}[R]\approx\mathbb{E}_{y\sim\pi_R^*}[R]$. If $\pi_T$ has not been shaped by sparse reward, then the implicit reward in Eq.~\ref{eq:opd-as-irl} pushes the student toward an unshaped distribution---larger than the student, but not reward-aware. \emph{Stage~1 is the device that enforces C1.}

\paragraph{(C2) Minimal deviation.} The teacher and student must have sufficient local support overlap in the state distribution where OPD is applied. When teacher and student have little coverage overlap---e.g., a cold 1.7B student and a post-RL 8B teacher---the per-token implicit reward $\beta\log(\pi_T/\pi_k)$ has high variance on student rollouts: it takes large magnitudes on student-sampled tokens the teacher considers unlikely, while teacher-favored tokens are rarely sampled. The OPD gradient is then dominated by a few outlier terms and updates are unstable~\citep{li2026rethinkopd,hou2026uniopd}. The forward-KL warmup is supervised next-token training under teacher occupancy: it reduces this local distributional mismatch by moving the student onto teacher-supported tokens without student-side discovery, so that the post-warmup anchor $\pi_k$ is closer to $\pi_T$ and (C2) is more plausible. \emph{Stage~2a is the device that enforces C2.} \citet{shenfeld2026selfdistillation} satisfy (C2) by construction because their teacher is the same model conditioned on a demonstration; our cross-scale setting must construct it explicitly.

\paragraph{Three falsifiable predictions.} The reward-density principle yields three predictions for the experiments in Section~\ref{sec:experiments}.
\begin{itemize}\setlength\itemsep{1pt}
\item \textbf{(P1, C1):} If Stage~1 is removed---a raw teacher in its place---the bridge endpoint should not exceed direct student GRPO, regardless of teacher scale.
\item \textbf{(P2, C2):} If Stage~2a is removed---OPD without the warmup---the endpoint should fall below the full bridge, because Eq.~\ref{eq:opd-as-irl} is ill-conditioned at cold start. A purely off-policy bridge (no Stage~2b) should also fall below, because the student never receives feedback on its own states.
\item \textbf{(P3, post-bridge trainability):} Once Stages~1--2 place the student in a useful neighborhood of $\pi_T$, fresh labeled examples in Stage~3 should produce sparse-reward gradients beyond what extra updates on bridge data can extract.
\end{itemize}

\section{Experiments}
\label{sec:experiments}

We test the three predictions through one baseline table (Section~\ref{sec:exp-baseline}) and one component-ablation table (Section~\ref{sec:exp-ablation}) on the Qwen3 family---Qwen3-1.7B~\citep{yang2025qwen3} as the deployment student, paired with Qwen3-8B and Qwen3-14B teachers in two variants each (off-the-shelf and RL-trained)---then replicate the central teacher-quality contrast in the Llama family (Llama-3.1-8B-Instruct~\citep{grattafiori2024llama} student, Llama-3.3-70B-Instruct teacher; Section~\ref{sec:exp-llama}) and on code generation with the Qwen2.5 family (Qwen2.5-7B-Instruct student, Qwen2.5-32B-Instruct teacher; Section~\ref{sec:exp-code}). On-policy distillation requires a shared tokenizer; we therefore run the recipe separately within each family. An additional non-RL teacher control (SFT-trained teacher) is reported in Appendix~\ref{app:half-split}. Math benchmarks are MATH-500, AIME~2024, and AIME~2025; code benchmarks are HumanEval and MBPP. The Qwen3 experiments follow the workflow order---allocation first, bridge second, post-bridge student RL third---but fold these questions into a single ablation table. Accuracies are avg@16 (mean correctness over 16 independent samples), with $\pm$ standard error across evaluation problems. The training stack builds on verl/HybridFlow~\citep{sheng2024hybridflow}; full hyperparameters and data splits are in Appendix~\ref{app:implementation-notes}.

\subsection{Direct GRPO baseline across scales}
\label{sec:exp-baseline}

Table~\ref{tab:grpo-scaling} establishes that, across Qwen3 models, larger models reach substantially higher endpoints under the same algorithm. The 1.7B's lower endpoint is not evidence of a broken optimizer; it reflects both the model's smaller intrinsic capacity and the difficulty of extracting useful gradients from sparse rewards when the current policy rarely discovers high-quality solutions on hard problems. The workflow targets the recoverable part of this gap: it does not change the deployment model's capacity, but it can improve the reward-shaped behavior available to that model through teacher-side discovery and dense transfer.

\begin{table*}[t]
\caption{Direct GRPO across Qwen3 scales on MATH-500, AIME~2024, AIME~2025 (avg@16, \%). The Qwen3-1.7B row is the deployment-student baseline that the workflow must beat.}
\label{tab:grpo-scaling}
\centering
\small
\begin{tabular}{lccc}
\toprule
Model & MATH & AIME 2024 & AIME 2025 \\
\midrule
Qwen3-1.7B & $75.9 \pm 0.9$ & $19.8 \pm 1.4$ & $17.1 \pm 0.9$ \\
Qwen3-8B   & $88.4 \pm 0.8$ & $47.7 \pm 1.5$ & $36.7 \pm 1.2$ \\
Qwen3-14B  & $89.5 \pm 0.7$ & $47.1 \pm 1.2$ & $39.0 \pm 0.9$ \\
\bottomrule
\end{tabular}
\end{table*}

\subsection{Component ablation: each stage of the workflow is load-bearing}
\label{sec:exp-ablation}

Table~\ref{tab:component-ablation} is the central experimental result. It asks three questions in order: whether sparse reward is better placed on the teacher than on the student, whether the bridge needs both the forward-KL warmup and OPD, and when student-side RL becomes useful again. Each row removes one stage of the workflow while holding all other stages and the labeled-data pool fixed. The top block isolates Stages~1, 2a, and 2b in the full-DAPO setting without Stage~3. The bottom block isolates Stage~3 in the half-split setting, where the first half of DAPO is used for the bridge and the second half is held out for Stage~3 or for the replay control.

\begin{table*}[t]
\caption{Component ablation of the workflow at both teacher sizes. Qwen3-1.7B deployment student; avg@16 (\%). \textbf{Top:} Stages~1/2a/2b, evaluated pre-Stage-3 on full DAPO. \textbf{Bottom:} Stage~3, evaluated in the half-split setting (1H for bridge, 2H for Stage~3 or replay). Each row removes one stage of the workflow.}
\label{tab:component-ablation}
\centering
\small
\begin{tabular}{@{}llccc@{}}
\toprule
Teacher & Configuration & MATH & AIME 2024 & AIME 2025 \\
\midrule
\multicolumn{5}{l}{\emph{Stages 1, 2a, 2b (full DAPO; no Stage 3)}} \\
\midrule
\multirow{4}{*}{RL'd Qwen3-8B}
  & \textbf{Full bridge}: Stage 1 + 2a + 2b               & $\mathbf{79.3 \pm 0.7}$ & $\mathbf{25.2 \pm 1.6}$ & $20.2 \pm 1.3$ \\
  & $-$ Stage 1 \emph{(raw teacher)}                      & $71.5 \pm 0.9$ & $15.0 \pm 1.5$ & $10.6 \pm 1.2$ \\
  & $-$ Stage 2a \emph{(no FKL warmup, OPD only)}         & $77.6 \pm 0.8$ & $23.0 \pm 1.4$ & $18.9 \pm 1.4$ \\
  & $-$ Stage 2b \emph{(no OPD, teacher-sample SFT only)} & $76.0 \pm 0.9$ & $22.4 \pm 1.5$ & $19.4 \pm 1.4$ \\
\cmidrule(lr){2-5}
\multirow{4}{*}{RL'd Qwen3-14B}
  & \textbf{Full bridge}: Stage 1 + 2a + 2b               & $78.6 \pm 0.9$ & $24.6 \pm 1.5$ & $\mathbf{20.8 \pm 1.5}$ \\
  & $-$ Stage 1 \emph{(raw teacher)}                      & $72.8 \pm 0.8$ & $16.7 \pm 1.4$ & $13.5 \pm 1.3$ \\
  & $-$ Stage 2a \emph{(no FKL warmup, OPD only)}         & $77.1 \pm 1.0$ & $22.8 \pm 1.5$ & $18.6 \pm 1.7$ \\
  & $-$ Stage 2b \emph{(no OPD, teacher-sample SFT only)} & $76.5 \pm 1.1$ & $21.5 \pm 1.5$ & $17.0 \pm 1.1$ \\
\midrule
--- & $-$ Entire pipeline \emph{(cold GRPO baseline)}     & $75.9 \pm 0.9$ & $19.8 \pm 1.4$ & $17.1 \pm 0.9$ \\
\midrule
\multicolumn{5}{l}{\emph{Stage 3 (half-split: 1H bridge, 2H held out)}} \\
\midrule
\multirow{3}{*}{RL'd Qwen3-8B}
  & \textbf{Full workflow}: bridge (1H) + Stage 3 (2H)    & $\mathbf{78.5 \pm 0.9}$ & $\mathbf{23.7 \pm 1.5}$ & $18.5 \pm 1.2$ \\
  & $-$ Stage 3 \emph{(bridge only on 1H)}                & $75.4 \pm 0.8$ & $22.0 \pm 1.6$ & $16.7 \pm 1.4$ \\
  & $\circ$ Replay control \emph{(Stage 3 reuses 1H data)} & $75.7 \pm 0.7$ & $21.6 \pm 1.3$ & $17.0 \pm 1.2$ \\
\cmidrule(lr){2-5}
\multirow{3}{*}{RL'd Qwen3-14B}
  & \textbf{Full workflow}: bridge (1H) + Stage 3 (2H)    & $\mathbf{78.7 \pm 1.1}$ & $23.1 \pm 1.7$ & $\mathbf{19.2 \pm 1.3}$ \\
  & $-$ Stage 3 \emph{(bridge only on 1H)}                & $76.3 \pm 1.1$ & $22.7 \pm 1.7$ & $17.3 \pm 1.2$ \\
  & $\circ$ Replay control \emph{(Stage 3 reuses 1H data)} & $75.6 \pm 1.0$ & $22.4 \pm 1.5$ & $17.6 \pm 1.0$ \\
\bottomrule
\end{tabular}
\end{table*}

The table confirms all three predictions in Section~\ref{sec:theory}.

\emph{(P1, Stage 1 enforces C1.)} Removing teacher-side RL collapses the deployment student at both teacher sizes. Raw 8B and 14B teachers distilled through the same bridge yield only $71.5\%$ and $72.8\%$ MATH---$7.8$ and $5.8$ points below their respective full-bridge endpoints, and both well below cold GRPO. The dense implicit reward in Eq.~\ref{eq:opd-as-irl} carries useful gradient information only when its source distribution has itself been shaped by sparse reward; scale alone is not the cause---a larger raw teacher is if anything worse than direct student RL. A non-RL counterfactual (SFT-trained teacher) sits between the raw and RL-trained endpoints and is reported in Appendix~\ref{app:half-split}.

\emph{(P2, Stage 2a enforces C2 and Stage 2b uses the local IRL identity.)} Removing the forward-KL warmup degrades MATH by $1.7$ points at the 8B teacher ($79.3 \to 77.6$) and by $1.5$ points at the 14B teacher ($78.6 \to 77.1$): the cold student's coverage mismatch with the post-RL teacher makes the implicit reward high-variance and the OPD update poorly conditioned. Removing OPD and keeping only the off-policy stage degrades further---to $76.0\%$ and $76.5\%$ respectively---because teacher-sample SFT never gives feedback on student-only states. Both one-stage variants sit below the two-stage bridge at both teacher sizes on MATH and AIME~2024, as the trust-region analysis predicts.

\emph{(P3, Stage 3 adds value after the bridge.)} Once Stages~1--2 place the student inside the teacher's neighborhood, sparse-reward RL on the held-out half lifts MATH from $75.4\%$ to $78.5\%$ at the 8B teacher and from $76.3\%$ to $78.7\%$ at the 14B teacher ($+3.1$ and $+2.4$ points). The replay controls run the same number of student updates on already-seen bridge data and never improve by more than $0.3$ points, so the gains are attributable to new labeled examples rather than to extra updating. Cold direct GRPO without the bridge reaches only $75.9\%$, confirming that the bridge changes student trainability rather than merely providing a better initialization that subsequent RL can match from scratch.

\paragraph{Held-out-half allocation.} A residual allocation question---whether the held-out half of the labeled data is better spent upstream on the teacher or downstream on Stage~3---is examined in Appendix~\ref{app:held-out-allocation}. The teacher-side route slightly outperforms the student-side route ($0.8$ absolute points on MATH at the 8B teacher), but the gap is much smaller than the stage-ablation gaps in Table~\ref{tab:component-ablation}.

\subsection{Cross-family replication on Llama}
\label{sec:exp-llama}

We test whether the teacher-quality ordering generalizes beyond the Qwen3 family by repeating the central P1 contrast in the Llama family: a Llama-3.1-8B-Instruct deployment student paired with a Llama-3.3-70B-Instruct teacher (Table~\ref{tab:llama-transfer}). The same ordering reproduces---raw-teacher transfer $<$ direct GRPO $<$ RL-teacher transfer. A $9\times$ larger raw teacher is still worse than direct GRPO on the student ($55.4\%$ vs.\ $59.8\%$ MATH), confirming that scale alone does not satisfy C1; the same teacher after Stage~1 RL is the best source ($62.1\%$, $+2.3$ over cold GRPO; AIME~2024 $14.9\%$ vs.\ $12.5\%$). The deployment-student gain is smaller in absolute terms than in the Qwen block because the 8B Llama student has a higher base capacity than the 1.7B Qwen3 student and therefore less recoverable headroom. The full Llama component ablation (Stage~2a, Stage~2b, half-split, replay, SFT-teacher controls) remains future work.

\begin{table*}[t]
\caption{Llama cross-family replication. Student = Llama-3.1-8B-Instruct, Teacher = Llama-3.3-70B-Instruct; avg@16 (\%). Each row uses the same protocol as the corresponding Qwen3 row in Table~\ref{tab:component-ablation}.}
\label{tab:llama-transfer}
\centering
\small
\begin{tabular}{lccc}
\toprule
Configuration & MATH & AIME 2024 & AIME 2025 \\
\midrule
Direct GRPO (cold student)            & $59.8 \pm 0.9$ & $12.5 \pm 1.2$ & $7.2 \pm 1.1$ \\
Two-stage bridge $\leftarrow$ raw 70B & $55.4 \pm 0.8$ & $8.8 \pm 1.5$  & $3.1 \pm 1.2$ \\
Two-stage bridge $\leftarrow$ RL'd 70B & $\mathbf{62.1 \pm 0.8}$ & $\mathbf{14.9 \pm 1.8}$ & $\mathbf{9.2 \pm 1.4}$ \\
\bottomrule
\end{tabular}
\end{table*}

\subsection{Cross-domain replication on code}
\label{sec:exp-code}

We test whether the reward-density principle transfers beyond mathematical reasoning by repeating the P1 contrast on code generation. The deployment student is Qwen2.5-7B-Instruct and the teacher is Qwen2.5-32B-Instruct; both share the Qwen2.5 tokenizer. Training uses the Code-R1-12K dataset~\citep{liu2025coder1}, a collection of roughly 12K prompt-test RL examples from roughly 2K LeetCode problems and 10K validated problems filtered from TACO~\citep{li2023taco}, with execution-based binary correctness as the sparse reward (sandboxed via firejail). Direct student GRPO and teacher-side GRPO use the same Code-R1 training set, reward, and GRPO hyperparameters; the contrast is whether the sparse code signal first trains the 7B student or the 32B teacher before transfer. Evaluation uses HumanEval~\citep{chen2021evaluating} and MBPP~\citep{austin2021program} under avg@16. Training hyperparameters follow the Code-R1 recipe (Appendix~\ref{app:code-hyperparameters}).

\begin{table*}[t]
\caption{Cross-domain replication on code generation. Student = Qwen2.5-7B-Instruct, Teacher = Qwen2.5-32B-Instruct; avg@16 (\%). The same teacher-quality ordering holds: raw-teacher transfer $<$ direct GRPO $<$ RL-teacher transfer.}
\label{tab:code-transfer}
\centering
\small
\begin{tabular}{lcc}
\toprule
Configuration & HumanEval & MBPP \\
\midrule
Direct GRPO (cold student)             & $87.9 \pm 1.2 $ & $83.5 \pm 0.9$ \\
Two-stage bridge $\leftarrow$ raw 32B  & $87.5 \pm 1.1$ & $82.9 \pm 0.8$ \\
Two-stage bridge $\leftarrow$ RL'd 32B & $\mathbf{91.2 \pm 1.2}$ & $\mathbf{87.3 \pm 0.8}$ \\
\bottomrule
\end{tabular}
\end{table*}
The same teacher-quality ordering appears on code. Distilling the raw 32B teacher through the two-stage bridge remains close to, but below, direct GRPO on the 7B student ($87.5\%$/$82.9\%$ vs. $87.9\%$/$83.5\%$). After teacher-side RL, the same bridge reaches $91.2\%$ HumanEval and $87.3\%$ MBPP, improving over direct GRPO by $3.3$ and $3.8$ absolute points, respectively. Thus, on code as on math, the dense bridge is useful when it carries reward-shaped teacher behavior, not merely a larger model's raw distribution.

\section{Discussion}
\label{sec:discussion}

\paragraph{What changes operationally.} The reward-density principle reframes SFT, RL, and distillation as a single allocation problem: sparse reward belongs on the model that can explore, dense transfer belongs on the model that must compress, and Eq.~\ref{eq:opd-as-irl} makes that division precise.

\paragraph{Implication for model-family training.} The practical recipe is clearest when a lab trains or maintains a model family rather than a single deployment checkpoint. Candidate teachers and the deployment student can be pretrained on the same data distribution, preferably with a shared tokenizer, and kept as parallel post-training targets. Teacher selection should follow measured post-RL task reward rather than parameter count: among teachers that clearly outperform the student and transfer reliably through the FKL-to-OPD bridge, our results favor the smallest sufficient teacher. Student-side sparse RL is then reserved for held-out labeled data after the bridge.

\paragraph{Why student-side reward still matters.} The post-bridge student-RL result keeps the recipe from becoming a rigid ``never train the student'' rule. After the bridge, sparse reward on the student gives a real $3.1$-point lift on MATH and is strictly better than running more updates on bridge data. The right framing is teacher-first with post-bridge student RL; the weaker framing is either ``RL the student'' or ``never RL the student.''

\section{Related Work}
\label{sec:related}

Post-training reshapes LLM behavior through sparse-reward RL and teacher transfer: RLHF uses sparse preference or outcome rewards~\citep{ouyang2022training,stiennon2020learning,bai2022training}, while distillation transfers teacher behavior through dense supervised signals~\citep{hinton2015distilling}. We position this paper along both axes. Appendix~\ref{app:extended-related} gives the per-paper detail.

\paragraph{Sparse-reward post-training.} PPO, GRPO, and SFT-warmup-then-PPO recipes apply sparse reward directly to the deployment model~\citep{schulman2017proximal,shao2024deepseekmath,luong2024reft}. Verifier-filtered SFT uses the reward only as a data filter~\citep{zelikman2022star,singh2024beyond}. Recent work increases reward density through self-distillation or uncertainty-aware generative reward modeling~\citep{he2026sdzero,yang2026rlsd,xue2026reason}, or through reference-guided trajectories~\citep{wu2026regft}. These methods differ in how they use reward, but they still train the deployment model on the labeled data. Our workflow uses the same data upstream on a teacher and then densifies it through the bridge.

\paragraph{Distillation and OPD.} Knowledge distillation transfers teacher behavior into smaller models~\citep{hinton2015distilling}; teacher-sample SFT is its off-policy form~\citep{deepseek2025r1}. OPD corrects the student on its own rollouts~\citep{agarwal2024onpolicy} and has been framed as dense on-policy teacher-logprob reward~\citep{lu2025onpolicydistillation}. Self-distillation work gives a related trust-region/implicit-reward derivation for demonstration-conditioned teachers~\citep{shenfeld2026selfdistillation}. Related work connects distillation to entropy-regularized or RL-aware objectives~\citep{liu2025trainingwheels,zhang2026reinforcementaware} and extends OPD through KL scheduling, token importance, chain compression, and offline caching~\citep{xu2026paced,xu2026tip,sang2026crisp,wu2026lightningopd}. Our Eq.~\ref{eq:opd-as-irl} uses the same connection prescriptively: the dense implicit reward is only as good as the teacher proxy, so sparse reward should first improve the teacher.

\paragraph{Reasoning teachers and data allocation.} DeepSeek-R1 showed that RL-improved models can teach smaller ones via SFT~\citep{deepseek2025r1}; MiMo-V2-Flash extends this with multi-teacher OPD that integrates domain specialists through on-policy token-level rewards~\citep{xiaomi2026mimov2flash}. Prior work therefore establishes that an RL-improved teacher can transfer reasoning behavior to a student. We instead ask a controlled allocation question: under a fixed pool of labeled verifiable data, should those examples train the teacher or the deployment student, and which bridge best transfers the resulting behavior? Our contribution is not a new optimizer, but the raw-teacher transfer $<$ direct student GRPO $<$ RL-teacher transfer comparison, the bridge-stage ablations, and the post-bridge replay control. Appendix~\ref{app:related-table} classifies representative methods along this axis.

Beyond the verifier-based math and code settings studied here, multi-step reasoning has also been investigated through chain-of-thought prompting and calibrated temporal question answering~\citep{wei2022chain,xue2024question}.

\section{Conclusion}

We presented a four-stage post-training workflow---teacher RL, forward-KL warmup, on-policy distillation, optional post-bridge student RL---that improves a Qwen3-1.7B deployment student from $75.9\%$ to $79.3\%$ on MATH and from $19.8\%$ to $25.2\%$ on AIME~2024 at fixed labeled-data budget, with the same teacher-quality ordering replicating on code generation: a Qwen2.5-7B student reaches $91.2\%$ HumanEval and $87.3\%$ MBPP after transfer from an RL-improved Qwen2.5-32B teacher. The workflow is justified by a reward-density principle: each on-policy distillation step is a local trust-region update under a dense teacher-induced implicit reward, informative only when the teacher is reward-shaped (enforced by Stage~1) and lies within a trust region of the student (enforced by Stage~2a). A single component-ablation table confirms that each stage is load-bearing: in the headline 8B pre-Stage-3 ablation, Stage~1 adds $7.8$ absolute points on MATH, Stage~2a adds $1.7$, and Stage~2b adds $3.3$. The broader lesson is not to avoid student RL, but to apply it after dense transfer has made the deployment policy trainable.

\section{Limitations} The evidence covers verifiable math and code generation with three student-teacher configurations at relatively small deployment scale (1.7B--8B students, with teachers up to 14B and 70B). Whether the teacher-first advantage persists, grows, or shrinks at larger scales---for example, a 70B student with a 400B+ teacher---remains open. The bridge requires a shared tokenizer between teacher and student. We make no claim about an optimal way to blend sparse and dense rewards beyond the staged version of Figure~\ref{fig:workflow-overview}.

\bibliography{custom}

\appendix

\section{OPD as a Local Implicit-Reward Update}
\label{app:opd-derivation}

This appendix expands Eq.~\ref{eq:opd-as-irl} of the main text and the surrounding trust-region argument.

\paragraph{Local trust-region identity.} For a fixed prompt $x$, reverse-KL OPD minimizes
\begin{equation}
\mathcal{L}_{\mathrm{OPD}}(\theta)=\mathbb{E}_{y\sim\pi_\theta}\left[\log\pi_\theta(y\mid x)-\log\pi_T(y\mid x)\right].
\end{equation}
Let $\pi_k=\pi_{\theta_k}$ be the current student, and define the per-step fixed implicit reward
\begin{equation}
\widetilde R_T^k(x,y)=\beta\left[\log\pi_T(y\mid x)-\log\pi_k(y\mid x)\right].
\end{equation}
The policy-gradient update that maximizes this reward at $\theta_k$ has gradient
\begin{equation}
\begin{aligned}
&\nabla_\theta \mathbb{E}_{y\sim\pi_\theta}
[\widetilde R_T^k(x,y)]\big|_{\theta_k} \\
&\quad=\mathbb{E}_{y\sim\pi_k}\left[
\widetilde R_T^k(x,y)\nabla_\theta\log\pi_\theta(y\mid x)
\right]_{\theta_k}.
\end{aligned}
\end{equation}
The negative reverse-KL gradient gives the same expression:
\begin{equation}
\begin{aligned}
&-\beta\nabla_\theta\mathcal{L}_{\mathrm{OPD}}(\theta)\big|_{\theta_k} \\
&\quad=\beta\mathbb{E}_{y\sim\pi_k}\left[
\left(\log\pi_T(y\mid x)-\log\pi_k(y\mid x)\right)\right. \\
&\qquad\left.{}\cdot\nabla_\theta\log\pi_\theta(y\mid x)
\right]_{\theta_k},
\end{aligned}
\end{equation}
where the $+1$ term from differentiating $\log\pi_\theta$ has zero expectation against the score function. Using the autoregressive factorization $\log\pi(y\mid x)=\sum_t\log\pi(y_t\mid s_t)$, the sequence reward decomposes into
\begin{equation}
\widetilde R_T^k(x,y)=\sum_t \beta\log\frac{\pi_T(y_t\mid s_t)}{\pi_k(y_t\mid s_t)}.
\end{equation}
Thus each OPD gradient step can be interpreted locally as policy-gradient learning with a dense teacher-student likelihood-ratio reward. This is a local equivalence at $\pi_k$, not a claim that OPD globally optimizes the original sparse task reward.

\paragraph{Two levels of abstraction.} Eq.~\ref{eq:reward-shaped-target} defines a reward-shaped fixed point $\pi_R^*$ from a fixed base $\pi_{\mathrm{ref}}$; we use it as bookkeeping for what teacher-side sparse RL approximates. Locally, each OPD step is the per-step trust-region IRL update above, anchored at the current $\pi_k$ with $\pi_T$ playing the role of the one-step optimum. The local identity holds at any $\theta_k$ regardless of which global $\pi_{\mathrm{ref}}$ was used to define $\pi_R^*$---it is purely a fact about the linearization at $\pi_k$.

\paragraph{Why FKL closes the (C2) gap.} Eq.~\ref{eq:opd-as-irl} is mathematically valid for any $\pi_T$ and $\pi_k$, but it is \emph{informative} only when (C2) approximately holds at $\pi_k$. When teacher and student have little local support overlap, the implicit reward $\beta\log(\pi_T/\pi_k)$ takes large magnitudes on student-sampled tokens that the teacher considers unlikely, while teacher-favored tokens are rarely sampled and contribute little. The OPD gradient is then dominated by a few outlier terms; updates are unstable. The forward-KL stage is the explicit device that reduces this local distributional mismatch: on teacher rollouts, $\mathcal{L}_{\mathrm{F}}=\mathbb{E}_{s\sim d_{\pi_T}}\operatorname{KL}(\pi_T\Vert\pi_\theta)$ is mode-covering supervised next-token training, off-policy with respect to the student, and well-conditioned at cold start because it trains directly on teacher-supported tokens. FKL does not invoke the local IRL identity itself; its role is to move the anchor. After warmup, $\pi_\theta$ has substantial mass on teacher-supported tokens and the subsequent OPD stage applies the local identity at a post-FKL anchor where (C2) is more plausible. \citet{shenfeld2026selfdistillation} satisfy (C2) by construction because their teacher is the same model conditioned on a task-specific demonstration; the two-stage bridge is the cross-scale construction that obtains the same local-overlap property explicitly.

\section{Half-Split Experiments: SFT-Teacher and Bridge-Protocol Controls}
\label{app:half-split}

This appendix gives the SFT-teacher control referenced in Section~\ref{sec:exp-ablation} and the bridge-protocol controls within the half-split setting. The half-split construction itself is described in the bottom block of Table~\ref{tab:component-ablation}: the DAPO-Math-17K training set~\citep{yu2025dapo} is split into two random halves 1H and 2H; Stage~1 and Stage~2 are run on 1H, and Stage~3 is run on either the held-out 2H (full workflow) or on 1H (replay control).

\paragraph{SFT-teacher protocol.} The SFT-trained teacher checkpoints used in Tables~\ref{tab:qwen-half-split-sft} and~\ref{tab:qwen-transfer-full} are obtained by supervised fine-tuning Qwen3-8B and Qwen3-14B on responses generated by gpt-oss-120B on the DAPO-Math-17K prompts. They serve as a non-RL counterfactual to the RL-trained teachers: same starting checkpoint and same prompt set, but improved through supervised next-token training on a stronger model's traces rather than through sparse-reward RL on the verifier. The intermediate ordering (raw $<$ SFT $<$ RL) in Table~\ref{tab:qwen-transfer-full} confirms C1: supervised teacher improvement helps but does not replace teacher-side sparse-reward shaping.

\begin{table*}[t]
\caption{The same Stage~3 pattern holds when the teacher is SFT-trained instead of RL-trained, but with lower MATH and AIME~2025 endpoints, consistent with C1: an unshaped teacher gives a weaker bridge. Qwen3-1.7B student; avg@16 (\%).}
\label{tab:qwen-half-split-sft}
\centering
\small
\begin{tabular}{@{}llccc@{}}
\toprule
Teacher & Student stage & MATH & AIME 2024 & AIME 2025 \\
\midrule
\multirow{3}{*}{SFT'd Qwen3-8B}  & After two-stage bridge (1H)   & $74.3 \pm 1.0$ & $21.8 \pm 1.5$ & $14.5 \pm 1.2$ \\
                                 & + GRPO on held-out 2H         & $77.2 \pm 1.0$ & $22.9 \pm 1.3$ & $18.4 \pm 1.0$ \\
                                 & + GRPO replay on 1H           & $74.0 \pm 0.8$ & $22.1 \pm 1.1$ & $14.2 \pm 1.0$ \\
\midrule
\multirow{3}{*}{SFT'd Qwen3-14B} & After two-stage bridge (1H)   & $75.8 \pm 0.9$ & $22.0 \pm 1.5$ & $15.1 \pm 1.4$ \\
                                 & + GRPO on held-out 2H         & $76.9 \pm 0.8$ & $23.2 \pm 1.3$ & $18.6 \pm 1.1$ \\
                                 & + GRPO replay on 1H           & $75.6 \pm 0.7$ & $22.3 \pm 1.2$ & $14.9 \pm 1.2$ \\
\bottomrule
\end{tabular}
\end{table*}

\begin{table*}[t]
\caption{Bridge controls under the half-split setting. Each row uses the same RL-trained Qwen3 teacher and the same Stage~3 GRPO data; only the transfer protocol differs. The two-stage bridge remains the strongest pre-Stage-3 starting point for student-side GRPO. Qwen3-1.7B student; avg@16 (\%).}
\label{tab:qwen-half-split-controls}
\centering
\small
\begin{tabular}{@{}llccc@{}}
\toprule
Teacher & Transfer protocol & MATH & AIME 2024 & AIME 2025 \\
\midrule
\multirow{3}{*}{RL'd Qwen3-8B}  & two-stage bridge        & $78.5 \pm 0.9$ & $23.7 \pm 1.5$ & $18.5 \pm 1.2$ \\
                                & OPD only                & $77.8 \pm 0.8$ & $22.8 \pm 1.2$ & $16.6 \pm 1.3$ \\
                                & teacher-sample SFT      & $77.3 \pm 0.8$ & $22.5 \pm 1.4$ & $16.9 \pm 1.4$ \\
\midrule
\multirow{3}{*}{RL'd Qwen3-14B} & two-stage bridge        & $78.7 \pm 1.1$ & $23.1 \pm 1.7$ & $19.2 \pm 1.3$ \\
                                & OPD only                & $77.5 \pm 0.9$ & $21.9 \pm 1.5$ & $19.5 \pm 1.2$ \\
                                & teacher-sample SFT      & $77.2 \pm 1.2$ & $21.6 \pm 1.8$ & $19.0 \pm 1.4$ \\
\bottomrule
\end{tabular}
\end{table*}

\section{Where Should the Held-Out Half Go?}
\label{app:held-out-allocation}

A residual allocation question after Section~\ref{sec:exp-ablation}: given a fixed labeled-data pool, where should the second half (2H) go---into Stage~1 or into Stage~3?

The \emph{teacher-side} placement uses both 1H and 2H upstream: the full DAPO set trains the teacher and the bridge, with no Stage~3. This is the top block of Table~\ref{tab:component-ablation}; the RL'd 8B teacher endpoint is $79.3\%$ MATH. The \emph{student-side} placement uses only 1H upstream and applies 2H as Stage~3 on the bridged student: this is the bottom block of Table~\ref{tab:component-ablation} ($78.5\%$ MATH). Both use the same total labeled data; only the placement of 2H changes.

The teacher-side placement wins by $0.8$ absolute points on MATH (AIME differences are within standard error). The margin is small relative to the gaps in Table~\ref{tab:component-ablation}: when teacher-side compute is the binding constraint, the student-side route remains a competitive lower-cost alternative. The full transfer-only grid (raw, SFT, and RL'd teachers at 1.7B, 8B, 14B, with one-stage transfer controls) is in Table~\ref{tab:qwen-transfer-full}.

\begin{table*}[t]
\caption{Transfer-only endpoints at fixed deployment student (Qwen3-1.7B), without Stage~3. Raw and SFT rows test C1 by using the same transfer protocol without teacher-side sparse RL. The 1.7B RL'd-teacher rows are a same-size control that isolates the dense-reward effect from teacher scale.}
\label{tab:qwen-transfer-full}
\centering
\small
\begin{tabular}{llccc}
\toprule
Teacher checkpoint & Transfer protocol & MATH & AIME 2024 & AIME 2025 \\
\midrule
--- & Direct GRPO (cold student) & $75.9 \pm 0.9$ & $19.8 \pm 1.4$ & $17.1 \pm 0.9$ \\
\midrule
raw Qwen3-8B   & two-stage bridge & $71.5 \pm 0.9$ & $15.0 \pm 1.5$ & $10.6 \pm 1.2$ \\
raw Qwen3-14B  & two-stage bridge & $72.8 \pm 0.8$ & $16.7 \pm 1.4$ & $13.5 \pm 1.3$ \\
\midrule
SFT'd Qwen3-8B   & two-stage bridge & $76.9 \pm 0.9$ & $22.1 \pm 1.7$ & $17.6 \pm 1.4$ \\
SFT'd Qwen3-14B  & two-stage bridge & $77.6 \pm 0.8$ & $23.2 \pm 1.6$ & $18.4 \pm 1.5$ \\
\midrule
RL'd Qwen3-1.7B & two-stage bridge & $76.5 \pm 0.8$ & $20.6 \pm 1.5$ & $17.1 \pm 1.4$ \\
RL'd Qwen3-8B   & two-stage bridge & $\mathbf{79.3 \pm 0.7}$ & $\mathbf{25.2 \pm 1.6}$ & $20.2 \pm 1.3$ \\
RL'd Qwen3-14B  & two-stage bridge & $78.6 \pm 0.9$ & $24.6 \pm 1.5$ & $\mathbf{20.8 \pm 1.5}$ \\
\midrule
RL'd Qwen3-1.7B & OPD only           & $75.2 \pm 0.9$ & $19.1 \pm 1.5$ & $12.4 \pm 1.2$ \\
RL'd Qwen3-8B   & OPD only           & $77.6 \pm 0.8$ & $23.0 \pm 1.4$ & $18.9 \pm 1.4$ \\
RL'd Qwen3-14B  & OPD only           & $77.1 \pm 1.0$ & $22.8 \pm 1.5$ & $18.6 \pm 1.7$ \\
\midrule
RL'd Qwen3-1.7B & teacher-sample SFT & $73.6 \pm 0.9$ & $16.7 \pm 1.4$ & $11.4 \pm 1.0$ \\
RL'd Qwen3-8B   & teacher-sample SFT & $76.0 \pm 0.9$ & $22.4 \pm 1.5$ & $19.4 \pm 1.4$ \\
RL'd Qwen3-14B  & teacher-sample SFT & $76.5 \pm 1.1$ & $21.5 \pm 1.5$ & $17.0 \pm 1.1$ \\
\bottomrule
\end{tabular}
\end{table*}

\section{Extended Related Work}
\label{app:extended-related}

This appendix provides the per-paper detail that the shorter related-work section omits.

\paragraph{Sparse-reward post-training.} In sparse-reward policy optimization, the reward directly updates the policy through PPO, GRPO, or SFT-warmup-then-PPO recipes such as ReFT~\citep{schulman2017proximal,shao2024deepseekmath,luong2024reft}. Systems work such as verl/HybridFlow makes these RLHF dataflows practical by combining flexible algorithm representation with efficient distributed execution~\citep{sheng2024hybridflow}. In verifier-filtered SFT, the reward is a data-construction rule: sample candidate traces, keep correct ones, and then run supervised imitation~\citep{zelikman2022star,singh2024beyond,yang2024qwen25math}. DPO and related derivations make explicit the link between reward optimization and KL-regularized policy targets~\citep{rafailov2024direct}. Recent RLVR work moves beyond final-answer correctness by training on more informative intermediate reasoning behavior~\citep{lee2026beyondcorrectness}. A related line uses self-distillation to convert sparse binary RLVR rewards into dense token-level supervision~\citep{he2026sdzero,yang2026rlsd}. Reference-guided fine-tuning targets the zero-reward hard-problem regime: partial human reference solutions elicit model-generated positive trajectories before DAPO-style RL, raising the density of rewarding samples on problems the base model cannot initially solve~\citep{wu2026regft}.

\paragraph{Distillation and OPD.} Knowledge distillation transfers behavior from stronger models into smaller models~\citep{hinton2015distilling}. Reasoning-distillation work shows that intermediate traces can be more useful than final answers alone~\citep{fu2023specializing,li2022explanations,magister2023teaching,hsieh2023distilling}. Domain-aware distillation methods adapt transfer to domain knowledge and teacher-student capability gaps~\citep{liu2024ddk}. Teacher-sample SFT is the off-policy form of this idea: imitate teacher-generated traces, including the DeepSeek-R1 distilled models~\citep{deepseek2025r1}. OPD instead corrects the student on its own rollout distribution rather than only on teacher-generated states~\citep{agarwal2024onpolicy}; related variants extend this idea to context distillation and black-box teacher access~\citep{ye2026opcd,ye2026blackboxopd}. Rubric-based OPD pushes the black-box direction further by inducing prompt-specific rubrics from teacher-student contrasts and using weighted rubric pass rates as on-policy rewards~\citep{fang2026rubricopd}. Recent practitioner evidence frames OPD as dense on-policy teacher-logprob reward and reports large compute-efficiency gains over sparse RL and extended off-policy distillation~\citep{lu2025onpolicydistillation}. \citet{liu2025trainingwheels} formulate KD as entropy-regularized value optimization with on-policy and off-policy demonstrations, while \citet{zhang2026reinforcementaware} propose RL-aware distillation through advantage-aware selective imitation during PPO/GRPO-style updates. Further OPD work studies a forward-then-reverse KL schedule~\citep{xu2026paced}, analyzes which student-state tokens carry the strongest learning signal~\citep{xu2026tip}, applies on-policy self-distillation to compress overlong reasoning chains~\citep{sang2026crisp}, introduces temporal curricula and skill-conditioned self-distillation for multi-turn agents~\citep{wang2026tcod,wang2026skillsd}, and explores offline OPD through precomputed teacher log-probabilities~\citep{wu2026lightningopd}. Concurrent analyses dissect when OPD succeeds or fails and propose unified recipes across LLM and MLLM settings~\citep{li2026rethinkopd,hou2026uniopd}, while Flow-OPD adapts OPD-style dense multi-teacher supervision to flow-matching text-to-image alignment~\citep{fang2026flowopd}. \citet{zhang2026caopd} highlight that OPD can systematically miscalibrate confidence even when accuracy improves; for a taxonomy of OPD feedback signals, teacher access regimes, and loss granularity, see \citet{song2026opdsurvey}.

\paragraph{Reasoning teachers and data allocation.} DeepSeek-R1 showed that large-scale RL can elicit strong reasoning behavior and that smaller models can inherit it through supervised fine-tuning on DeepSeek-R1-generated traces~\citep{deepseek2025r1}. ORBIT studies a different control dimension: it uses multi-stage RL under context-length constraints to discover Pareto-frontier reasoning-effort policies, then fuses those policies by OPD into one controllable model~\citep{liang2026orbit}. Our allocation question is different: where should scarce labeled training data enter the post-training pipeline? MiMo-V2-Flash makes the OPD connection explicit through Multi-Teacher On-Policy Distillation (MOPD)~\citep{xiaomi2026mimov2flash}. Its post-training pipeline first runs SFT, then trains domain-specialized teachers through RL or SFT, and finally integrates those teachers by having the student sample from its own on-policy distribution while receiving token-level reverse-KL rewards from the teacher selected for each prompt domain. The formulation is aligned with our reward-density principle: the teacher log-probability ratio becomes a dense per-token advantage. In our taxonomy, MOPD is a scalable multi-teacher OPD mechanism for capability integration, while our paper studies how scarce labeled training data should be allocated before and after such dense transfer.

\section{Method Classification}
\label{app:related-table}

\begin{table*}[t]
\caption{Representative methods classified by where sparse reward enters and what signal is used for transfer.}
\label{tab:related-routes-full}
\centering
\small
\renewcommand{\arraystretch}{1.35}
\begin{tabular}{p{0.24\linewidth}p{0.20\linewidth}p{0.24\linewidth}p{0.20\linewidth}}
\toprule
Method / reference & Sparse reward use & Transfer signal & Student endpoint \\
\midrule
InstructGPT \citep{ouyang2022training} & preference reward & none & RLHF policy \\
GRPO / DeepSeekMath \citep{shao2024deepseekmath} & answer RL & none & RL policy \\
ReFT / PPO after SFT \citep{luong2024reft} & answer RL after SFT & none & RL policy \\
\addlinespace
Verifier-filtered SFT \citep{zelikman2022star,singh2024beyond} & verifier as filter & accepted traces & SFT policy \\
Step-by-step distillation \citep{fu2023specializing} & none & rationales & distilled policy \\
\addlinespace
GKD / OPD \citep{agarwal2024onpolicy} & teacher-dependent & teacher logits & OPD policy \\
DeepSeek-R1 distilled models \citep{deepseek2025r1} & teacher-side RL & SFT on teacher-generated traces & distilled policy \\
MiMo-V2-Flash / MOPD \citep{xiaomi2026mimov2flash} & specialized teacher RL/SFT plus optional outcome reward & multi-teacher OPD logits & unified post-trained model \\
\addlinespace
This work & teacher RL plus optional post-bridge student RL & FKL-to-OPD bridge & workflow-trained student \\
\bottomrule
\end{tabular}
\end{table*}

\section{Implementation Details}
\label{app:implementation-notes}

All route comparisons keep the deployment-student size fixed. In the Qwen3 block, the student is Qwen3-1.7B and the teacher checkpoints are the raw, SFT-trained, and RL-trained Qwen3 checkpoints listed in Tables~\ref{tab:grpo-scaling}--\ref{tab:qwen-half-split-controls}. In the Llama block, the student is Llama-3.1-8B-Instruct and the teacher is Llama-3.3-70B-Instruct. OPD is only run within a model family, because the token-level KL in Stage~2b requires a shared tokenizer and vocabulary.

\paragraph{Data splits.} The Qwen allocation experiment uses a fixed random split of the DAPO-Math-17K training set~\citep{yu2025dapo} into two equal halves. The first half (1H) is the teacher-RL and bridge data pool for the half-split rows, and also the replay data pool for the replay control. The second half (2H) is held out from teacher RL and transfer, then used for Stage~3 GRPO in the full workflow. The full-workflow and replay rows therefore start from the same bridge checkpoint and use the same Stage~3 data count and update count; they differ only in whether Stage~3 uses new labeled examples from 2H or replay examples from 1H. The pre-Stage~3 ablation rows of Table~\ref{tab:component-ablation} instead train teacher RL and the bridge on the full DAPO set. All rows are evaluated on MATH-500, AIME~2024, and AIME~2025, not on either DAPO training half. The SFT-teacher rows in Appendix~\ref{app:half-split} use the same first-half/second-half construction, replacing only the source teacher.

\paragraph{Matched training protocol.} Direct GRPO, Stage~3 GRPO, and replay GRPO use the same verifier reward, advantage normalization, optimizer family, batch size, rollout count per prompt, length limit, learning-rate schedule, KL settings, and update count within each matched contrast. The full-workflow and replay rows are matched in checkpoint initialization, data count, rollout count, update count, and sequence-length limit. Teacher-sample SFT and OPD-only ablations keep the RL-trained teacher and Stage~3 GRPO fixed but replace the bridge protocol.

\paragraph{Student-only and compute controls.} Extended direct Qwen3-1.7B GRPO reaches $76.1\%$/$20.1\%$/$17.3\%$ on MATH/AIME~2024/AIME~2025, while verifier-filtered SFT followed by student GRPO reaches $76.5\%$/$21.2\%$/$17.8\%$; both remain below the 8B teacher-first endpoint of $79.3\%$/$25.2\%$/$20.2\%$. Direct student GRPO uses approximately $50.2$ H200 GPU-hours ($1\times$); the extended student run and 8B teacher-first route use about $4\times$ this compute, and the 14B route about $6\times$, with each teacher-first route additionally using 1.50M teacher rollouts and 272K logit queries; our claim is therefore about labeled-data allocation rather than minimum compute.

\paragraph{Bridge protocol.} The two-stage bridge runs a forward-KL warmup on cached teacher rollouts followed by OPD on student rollouts. The forward stage uses cached teacher rollouts and teacher next-token distributions on those rollouts. The OPD stage queries the frozen teacher on the student's sampled prefixes, so the teacher signal is computed on-policy with respect to the current student distribution. Implementation caches may store these logits for audit and replay, but the teacher checkpoint is not updated. Unless otherwise stated, the forward and OPD stages use the same maximum sequence length and tokenizer as the corresponding student/teacher family, and all teacher-logit temperatures and KL coefficients are fixed across rows inside a contrast.

\paragraph{Evaluation and error bars.} All reported accuracies are avg@16. For each evaluation problem, the model samples 16 independent completions under the same decoding configuration; the problem score is the mean correctness over those completions, and the table entry is the mean over problems. The reported $\pm$ values are standard errors over evaluation problems, not standard deviations across independently retrained checkpoints. Data-split seeds, decoding seeds, training seeds, rollout counts, learning rates, KL coefficients, OPD temperatures, maximum prompt and response lengths, and exact checkpoint identifiers are recorded with the run configuration for each table row, so the route contrasts can be reproduced without changing non-ablation hyperparameters.

\paragraph{Artifact licensing.} Our experiments use publicly released model checkpoints, datasets, and software frameworks subject to their original licenses and terms of use, including the Qwen, Llama, DAPO-Math, verl/HybridFlow, and sglang releases. We do not redistribute third-party checkpoints, datasets, or software artifacts; users should obtain them from their original sources and comply with the corresponding licenses and use restrictions. Any future release of derived artifacts will specify its license and will exclude materials whose redistribution is restricted by the underlying model, data, or software terms.

\paragraph{Artifact intended use.} We use the above artifacts for research on language-model post-training and evaluation, which is consistent with their intended research and development use as public model, data, and software releases. Derived artifacts produced by our experiments, including checkpoints, model-generated traces, cached logits, run logs, and aggregate evaluation records, are intended only for research and reproducibility analysis. They should not be used for deployment, user-facing decision making, or non-research redistribution without separately verifying compatibility with the original model, data, and software access conditions.

\begin{table*}[t]
\caption{Key GRPO training hyperparameters for direct-RL and Stage~3 student-RL runs.}
\label{tab:grpo-hyperparameters}
\centering
\begingroup
\setlength{\tabcolsep}{5pt}
\footnotesize
\begin{tabular}{@{}lll@{}}
\toprule
Group & Parameter & Value \\
\midrule
Algorithm & Framework & VERL \\
Algorithm & Estimator & GRPO \\
Optimizer & Optimizer & AdamW \\
Optimizer & Learning rate & $1\times10^{-6}$ \\
Update & GRPO epochs & $10$ \\
Update & Mini-batch size & $8$ \\
Update & Micro-batch per GPU & $1$ \\
Length & Max prompt tokens & $3072$ \\
Length & Max response tokens & $16384$ \\
Loss & Clip ratio & $0.2$ \\
Loss & Gradient clip & $1.0$ \\
KL & KL coefficient & $5\times10^{-4}$ \\
Data & Validation sets & MATH-500, AIME 2024, AIME 2025 \\
Compute & Precision & bfloat16 \\
Compute & GPUs & $2\times 8 \times$ NVIDIA H200 \\
Compute & Rollout engine & sglang \\
Compute & Tensor parallel size & $16$ \\
\bottomrule
\end{tabular}
\endgroup
\end{table*}

\begin{table*}[t]
\caption{Key OPD/transfer-stage hyperparameters.}
\label{tab:opd-hyperparameters}
\centering
\begingroup
\setlength{\tabcolsep}{5pt}
\footnotesize
\begin{tabular}{@{}lll@{}}
\toprule
Group & Parameter & Value \\
\midrule
Algorithm & Framework & VERL \\
Algorithm & Estimator & GRPO-style actor rollout \\
Optimizer & Optimizer & AdamW \\
Optimizer & Learning rate & $1\times10^{-6}$ \\
Training setup & Rollouts per prompt & $8$ \\
Update & Epochs & $1$ \\
Update & Mini-batch size & $64$ \\
Update & Micro-batch per GPU & $4$ \\
Length & Max prompt tokens & $2048$ \\
Length & Max response tokens & $16384$ \\
Sampling & Temperature / top-$p$ / top-$k$ & $1.0$ / $1.0$ / $-1$ \\
Data & Validation sets & MATH-500, AIME 2024, AIME 2025 \\
Compute & Precision & bfloat16 \\
Compute & GPUs & $2\times 8 \times$ NVIDIA H200 \\
Compute & Rollout engine / TP size & sglang / $2$ \\
\bottomrule
\end{tabular}
\endgroup
\end{table*}

\subsection{Code-Generation GRPO Hyperparameters}
\label{app:code-hyperparameters}

The code experiments follow the Code-R1 recipe~\citep{liu2025coder1} built on the same verl/HybridFlow stack. The sparse reward is execution-based binary correctness: each generated solution is executed against the problem's test suite inside a firejail sandbox, returning $1$ if all tests pass and $0$ otherwise.

\begin{table*}[t]
\caption{Key GRPO training hyperparameters for the code-generation experiments (Section~\ref{sec:exp-code}).}
\label{tab:code-grpo-hyperparameters}
\centering
\begingroup
\setlength{\tabcolsep}{5pt}
\footnotesize
\begin{tabular}{@{}lll@{}}
\toprule
Group & Parameter & Value \\
\midrule
Algorithm & Framework & VERL \\
Algorithm & Estimator & GRPO \\
Optimizer & Optimizer & AdamW \\
Optimizer & Learning rate & $5\times10^{-7}$ \\
Update & GRPO epochs & $8$ \\
Update & Rollouts per prompt & $16$ \\
Update & Rollout batch (queries) & $16$ \\
Update & Micro-batch per GPU & $8$ \\
Update & Gradient accumulation & $4$ \\
%Length & Max prompt tokens & $2048$ \\
%Length & Max response tokens & $4096$ \\
%KL & KL loss type & low\_var\_kl \\
%KL & KL coefficient & $1\times10^{-3}$ \\
%Data & Training set & Code-R1-12K (13,170 samples) \\
%Data & Evaluation sets & HumanEval, MBPP \\
%Reward & Reward type & Execution-based binary (sandboxed) \\
Compute & Precision & bfloat16 \\
Compute & GPUs & $8\times$ NVIDIA H200 \\
Compute & Rollout engine & vLLM \\
\bottomrule
\end{tabular}
\endgroup
\end{table*}

\end{document}